\documentclass[letterpaper]{article} % DO NOT CHANGE THIS
\usepackage{aaai25}  % DO NOT CHANGE THIS
\usepackage{times}  % DO NOT CHANGE THIS
\usepackage{helvet}  % DO NOT CHANGE THIS
\usepackage{courier}  % DO NOT CHANGE THIS
\usepackage[hyphens]{url}  % DO NOT CHANGE THIS
\usepackage{graphicx} % DO NOT CHANGE THIS
\usepackage{natbib}  % DO NOT CHANGE THIS AND DO NOT ADD ANY OPTIONS TO IT
\usepackage{caption} % DO NOT CHANGE THIS AND DO NOT ADD ANY OPTIONS TO IT
\usepackage{multirow}
\usepackage{colortbl}
\usepackage{xcolor}
\usepackage{graphicx} % for \resizebox
\usepackage[utf8]{inputenc}
\usepackage{amsmath} % For math fonts, symbols and environments
\usepackage{algorithm} % For the algorithm environment
\usepackage[utf8]{inputenc}
\usepackage{amssymb}
\usepackage{mathtools}
\usepackage{amsthm}
\usepackage{multirow}
\usepackage{diagbox}
\usepackage{bbding}
\usepackage{bm}
\usepackage{adjustbox}
\usepackage{graphicx}
\usepackage{multirow}
\usepackage{makecell}
\usepackage[capitalize,noabbrev]{cleveref}
\usepackage{microtype}
\usepackage{graphicx}
\usepackage{subfigure}
\usepackage{booktabs}
\usepackage{colortbl}
\usepackage{siunitx} % Required for alignment
\usepackage{booktabs} % For prettier tables
\usepackage{tabularx}
\usepackage{amssymb}
\usepackage{pifont}

\usepackage{algorithm}
\usepackage{algorithmic}

\usepackage{newfloat}
\usepackage{listings}
\DeclareCaptionStyle{ruled}{labelfont=normalfont,labelsep=colon,strut=off} % DO NOT CHANGE THIS
\floatstyle{ruled}
\newfloat{listing}{tb}{lst}{}
\floatname{listing}{Listing}
\title{BUFF: Bayesian Uncertainty Guided Diffusion Probabilistic Model for Single Image Super-Resolution}
\author{
    Zihao He\textsuperscript{\rm 1},
    Shengchuan Zhang\textsuperscript{\rm 1},
    Runze Hu\textsuperscript{\rm 2},
    Yunhang Shen\textsuperscript{\rm 3},
    Yan Zhang\textsuperscript{\rm 1,}\thanks{Corresponding Author.}\\
}
\affiliations{
    \textsuperscript{\rm 1}Key Laboratory of Multimedia Trusted Perception and Efficient Computing, Ministry of Education of China, Xiamen University, 361005, P.R. China.\\
    \textsuperscript{\rm 2}Department of Information Science, Tsinghua university.\\
    \textsuperscript{\rm 3}Tencent Youtu Lab, Shanghai 200233, China.
    36920221153081@stu.xmu.edu.cn, \{zsc$\_$2016, bzhy986\}@xmu.edu.cn\\
    \vspace{8mm}
}

\usepackage{bibentry}
\begin{document}

\maketitle

\begin{abstract}
Super-resolution (SR) techniques are critical for enhancing image quality, particularly in scenarios where high-resolution imagery is essential yet limited by hardware constraints. Existing diffusion models for SR have relied predominantly on Gaussian models for noise generation, which often fall short when dealing with the complex and variable texture inherent in natural scenes. To address these deficiencies, we introduce the Bayesian Uncertainty Guided Diffusion Probabilistic Model (BUFF). BUFF distinguishes itself by incorporating a Bayesian network to generate high-resolution uncertainty masks.
These masks guide the diffusion process, allowing for the adjustment of noise intensity in a manner that is both context-aware and adaptive. This novel approach not only enhances the fidelity of super-resolved images to their original high-resolution counterparts but also significantly mitigates artifacts and blurring in areas characterized by complex textures and fine details. The model demonstrates exceptional robustness against complex noise patterns and showcases superior adaptability in handling textures and edges within images. Empirical evidence, supported by visual results, illustrates the model's robustness, especially in challenging scenarios, and its effectiveness in addressing common SR issues such as blurring. Experimental evaluations conducted on the DIV2K dataset reveal that BUFF achieves a notable improvement, with a +0.61 increase compared to baseline in SSIM on BSD100, surpassing traditional diffusion approaches by an average additional +0.20dB PSNR gain. These findings underscore the potential of Bayesian methods in enhancing diffusion processes for SR, paving the way for future advancements in the field.
\end{abstract}

\section{Introduction}
Image super-resolution (SR), the art and science of enhancing the resolution of images, has undergone a remarkable evolution over the past decades. From its inception, where basic interpolation techniques like bicubic and Lanczos resampling were the norm, to the advent of deep learning, which has radically transformed the landscape, SR has been pivotal in fields ranging from satellite imaging and surveillance to medical imaging and entertainment.

\begin{figure}
    \centering
    \includegraphics[width=\linewidth,height=0.7\linewidth]{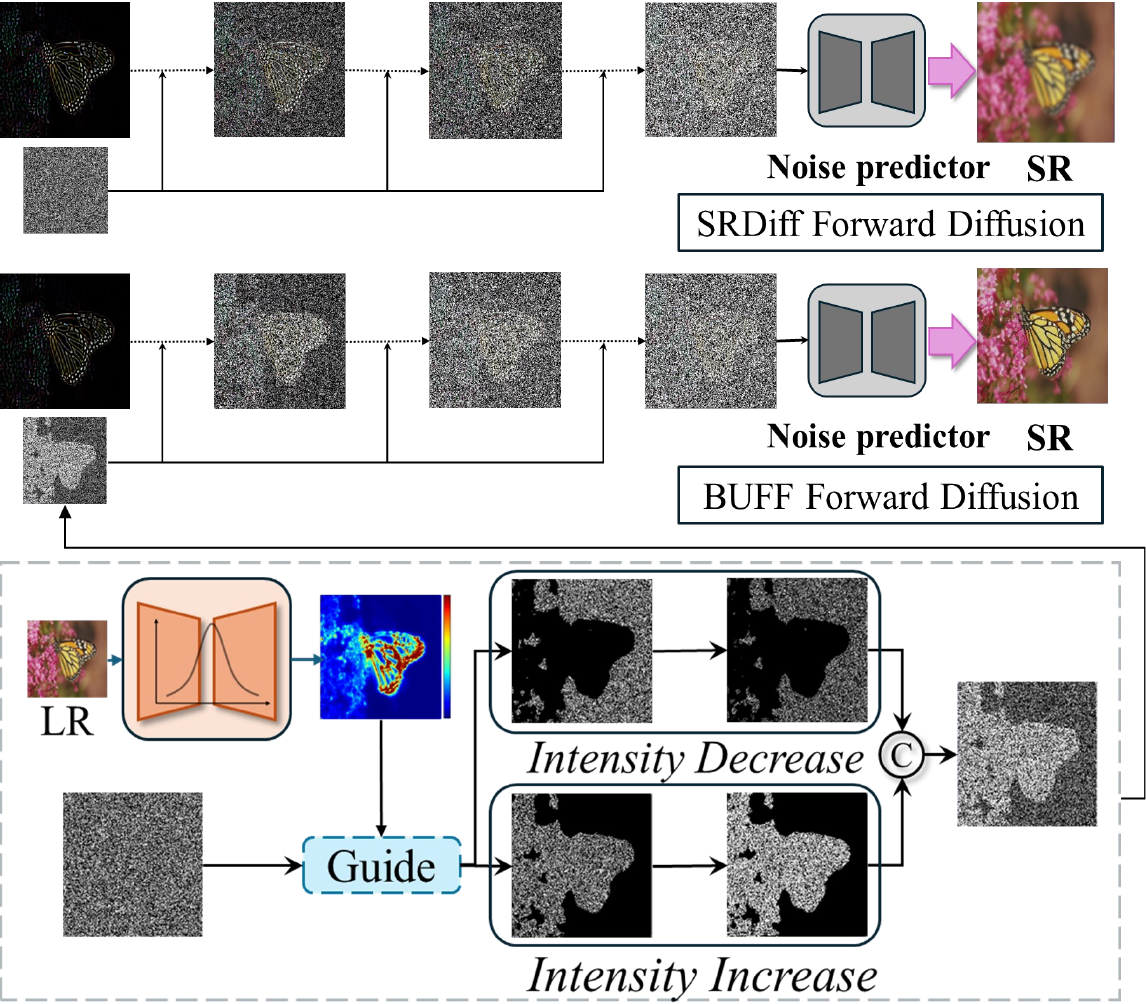}
    \caption{Comparison of the noise addition strategies in the forward diffusion phase between SRDiff\cite{srdiff} and BUFF approaches, with BUFF utilizing a Bayesian model to generate uncertainty masks that guide noise intensity adjustments across different regions of the image for SR results.}
    \label{fig:cover}
    \vspace{-1mm}
\end{figure}

The breakthrough comes with the introduction of Convolutional Neural Networks (CNNs) into the SR domain. Early models such as SRCNN \cite{srcnn} opened the floodgates to a new era of deep learning-based SR, offering significant improvements over traditional methods. These CNN-based models \cite{san,rcan,edsr} excel in mapping low-resolution (LR) to high-resolution (HR) images, learning a direct end-to-end transformation. Despite their success, they often fall short in capturing the perceptual nuances of images \cite{survey1,survey2,survey4}, leading to outcomes that, while technically accurate, lack the rich textures and details that make images lifelike \cite{survey3,he}. Transformers, known for their prowess in handling sequential data, make their mark on image SR with their ability to capture long-range dependencies within an image. By focusing on global information, Transformer-based SR models such as \cite{transformer2,hat,swinir} offer a promising approach to reconstructing images. Yet, their computational intensity and substantial memory footprint make them less viable for real-time applications, limiting their widespread adoption \cite{transformer1}. Generative Adversarial Networks (GANs) shifted the focus towards generating images that not only are high resolution but also perceptually convincing \cite{gansurvey1,survey4}. By employing a dual network system, one to generate images and another to critique them, GAN-based SR models like ESRGAN\cite{esrgan},BebyGAN\cite{bebygan} and Real-ESRGAN\cite{realesrgan} have pushed the boundaries of what's possible in terms of image quality. However, the adversarial nature of these models can lead to instability during training, sometimes resulting in artifacts or overly stylized images that detract from the realism of the output \cite{survey5}.

Diffusion-based super-resolution (SR) models, such as those referenced in \cite{srdiff,acdmsr,resdiff,samdiff,idm}, are at the forefront of image reconstruction, showcasing impressive capabilities in enhancing image quality. These models simulate the natural diffusion process through iterative noise introduction and attenuation to refine image details. However, despite their effectiveness in rendering rich textures, they assume noise is independent and identically distributed \cite{srdiff,acdmsr}, potentially overlooking the distinct data distributions across different image regions. This can compromise structural integrity, leading to inconsistent textures and noticeable artifacts. While \cite{samdiff} attempts to address this by modulating Gaussian noise with a SAM-generated mask \cite{sam}, it still struggles with precise pixel restoration, particularly along edges. Additionally, managing noise in diffusion processes, though innovative, introduces significant computational burdens and parameter sensitivity, posing challenges for practical deployment.

In this context, Bayesian models offer a promising complementary perspective. Known for their role in probabilistic image generation, Bayesian methods \cite{bsurvey1,bsurvey2} facilitate image creation and enable the quantification of uncertainty in predictions \cite{bayesiansr1}. Despite the potential synergy, integrating Bayesian approaches with diffusion-based generative models remains largely unexplored. While diffusion models excel at transforming noise into coherent structures, they typically lack a formalized method to evaluate or leverage the inherent uncertainty of this process. Incorporating a Bayesian framework into diffusion SR models could enhance both the fidelity and precision of synthesized images, particularly in regions of high uncertainty \cite{bayesiansr2,bdiff1}. However, directly integrating Bayesian principles into diffusion models is challenging due to the computational intensity and instability of Bayesian training, compounded by the difficulty of achieving convergence while maintaining the accuracy of the noise predictor.

In response to these challenges,depicted in Figure.\ref{fig:cover}, this paper introduces a novel framework, BUFF (Bayesian Uncertainty Guided Diffusion Model). Drawing on the structure of the \cite{srdiff} diffusion model, BUFF ingeniously modifies the noise distribution by embedding a Bayesian model to guide the noise prediction with pixel-level uncertainty insights. By doing so, BUFF injects structural information into the diffusion process more judiciously, which is critical for advancing the super-resolution performance of the diffusion model. Specifically, for each low-resolution (LR) image in the training set, our Bayesian model generates a corresponding uncertainty map for the high-resolution (HR) version. This map, delineating areas of high and low uncertainty, undergoes a refinement process that scales the uncertainties with designated multipliers, resulting in a modulation mask. During training, this mask directs the diffusion model's attention, specifically the noise predictor (U-Net), to areas of higher uncertainty, ensuring that these regions receive heightened focus. The modulated noise, obtained by applying the modulation mask to Gaussian noise, is then used alongside the LR image as part of the conditional inputs to the model. The masks for the training samples are pre-generated by the Bayesian model and can be reused, enhancing efficiency across training epochs. Through extensive experimentation on various standard image SR benchmarks, BUFF has been shown to surpass current diffusion-based methodologies. Our method's versatility is further demonstrated through additional experiments in tasks such as deblurring and face super-resolution, confirming the scalability of BUFF and its practicality in diverse multimedia applications. 

\section{Related Work}

\subsection{Uncertainty in Bayesian Deep Learning}
The incorporation of uncertainty into deep learning, especially within the Bayesian framework, has gained substantial attention for its potential to enhance model robustness and interpretability \cite{bsurvey1}. Bayesian Deep Learning(BDL) offers a principled approach to quantify uncertainty, distinguishing between aleatoric (data) and epistemic (model) uncertainties \cite{bsurvey5}. Pioneering research by \cite{bsurvey2} and \cite{bsurvey1} has shown how BDL can be applied across various tasks, including vision \cite{ddfm} and natural language processing \cite{bsentiment}, to improve decision-making under uncertainty. In the context of SR, understanding and modeling uncertainty can significantly enhance the quality of reconstructed images \cite{bayescap,bayesiansr1,bayesiansr2}. By identifying areas of high uncertainty, models can adaptively refine these regions, potentially leading to higher quality reconstructions \cite{bayesiansr3}. Recent works, such as those by \cite{bmri1,bmri2}, have explored the application of BDL in medical image reconstruction, demonstrating how uncertainty can guide the recovery of fine details while assessing the confidence in the model's predictions. This research underscores the value of incorporating uncertainty into SR models, suggesting avenues for improving SR techniques through a better understanding of where models are most and least certain in their predictions.

\begin{figure*}[htbp]
    \centering
    \includegraphics[width=0.9\linewidth,height = 0.55\linewidth]{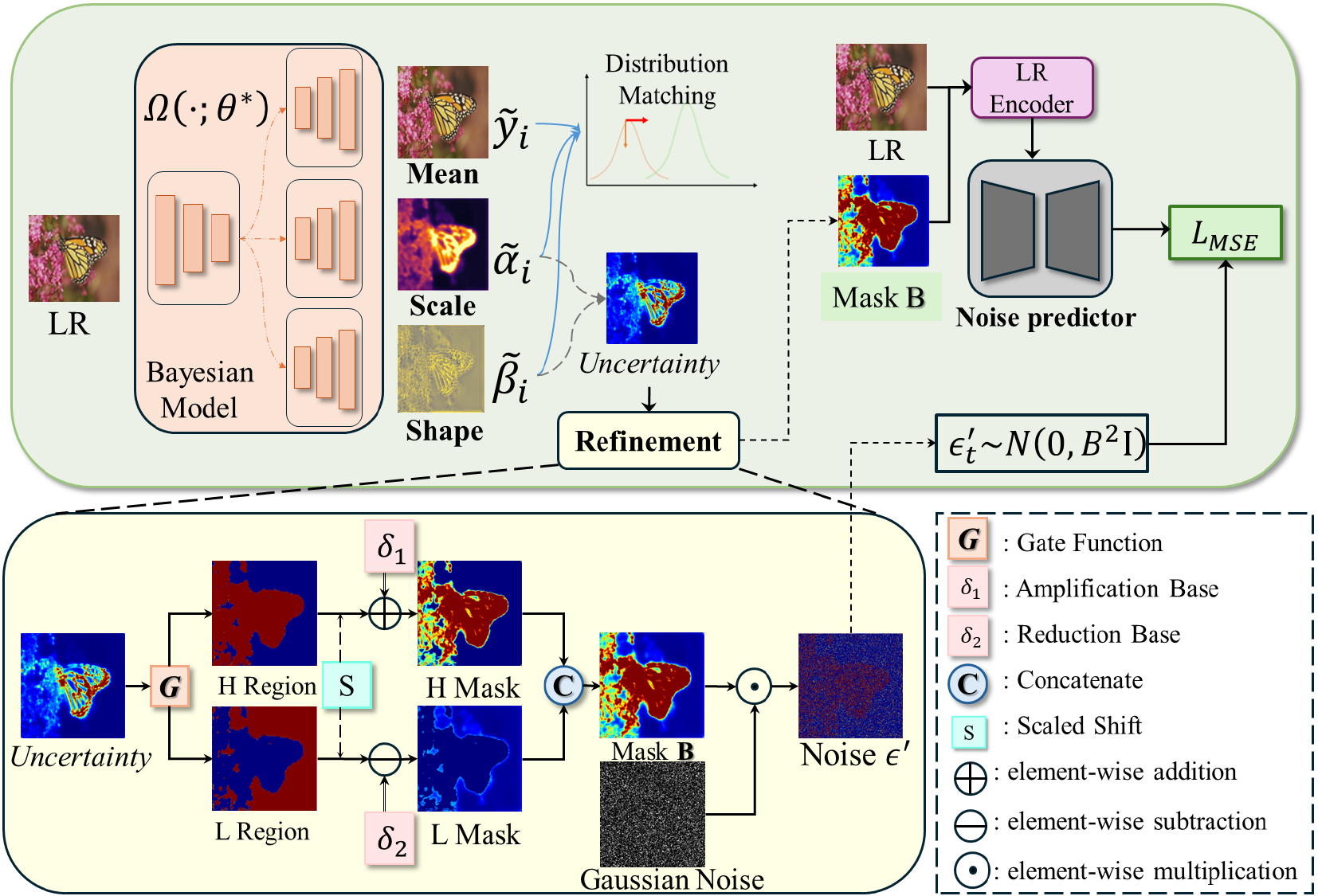}
    \caption{Process diagram showcasing the Bayesian modeling for uncertainty estimation in image super-resolution, where a Bayesian neural network refines uncertainty measures to guide noise modulation, enhancing the LR to HR reconstruction process.}
    \label{fig:arch}
\end{figure*}   

\subsection{Diffusion-Based Super Resolution}
Diffusion-based models mark an innovative paradigm in generative modeling, closely emulating the ebb and flow of noise addition and subtraction to fabricate or restore images. These models derive their methodology from the physical diffusion process, involving a forward phase of progressive noise application and a backward phase aiming to revert the noised data to its original form \cite{diffsurvey1}. Across various domains, diffusion methodologies, as pioneered by \cite{noise1,noise2}, stand out as a robust alternative to conventional generative techniques, facilitating the synthesis of high-fidelity images by the meticulous orchestration of noise. Distinct from GANs, which are prone to generating synthetic artifacts \cite{esrgan}, diffusion models are celebrated for their aptitude in crafting images replete with realistic textures and finer details \cite{diffsurvey3}. Despite their promise, the computational demands and the intricate optimization required for specific applications, such as SR tasks, present substantial challenges. Nevertheless, recent innovations that meld diffusion models with supplementary architectures have made strides in enhancing training efficiency and elevating SR performance. Pioneering work by \cite{srdiff,samdiff,resdiff} has underscored the prowess of diffusion-based models in the SR arena, setting new benchmarks in image clarity and lifelikeness. This burgeoning research underpins a keen and expanding interest in diffusion-based strategies as a potent mechanism for image reconstruction, particularly in the realm of SR tasks.

\section{Method}
\label{sec:method}
In this study, we present an innovative Bayesian-guided diffusion process specifically designed for image super-resolution (SR). This method enhances traditional diffusion-based SR models by integrating a Bayesian network’s uncertainty estimates directly into the diffusion sequence, tailoring noise addition from each input image based on per-pixel uncertainty. This allows for a stochastic process that adapts to the image's inherent confidence levels, facilitating a more intelligent and dynamic reconstruction approach.

\textbf{Bayesian Training and Inference for Uncertainty Estimation in SR.} The key component of our model is the estimation of uncertainty for each pixel in low-resolution (LR) images. This estimation is grounded in Bayesian inference, which provides a robust mechanism for assessing confidence and managing the inherent ambiguities in the SR process. Beforehand, based on \cite{bayescap}, we discuss a model initialized from scratch (training the network from an uninitialized state rather than using pretrained weights) to tackle the target task and estimate uncertainty, denoted as $\boldsymbol{\Psi}s(\cdot;\zeta):\mathbb{R}^m\to\mathbb{R}^n$, with $\zeta$ being its trainable parameters and $\boldsymbol{\Psi}$ refers to the set of parameters modeling the uncertainty distribution. This model aims to estimate the parameters of the output distribution $\mathcal{P}_{Y|X}$ (representing the predicted probability distribution over possible super-resolved outputs) to account for aleatoric uncertainty. For an input $x_i$, it generates parameters ${\hat{\mathbf{y}}_i,\hat{\nu}i\ldots\hat{\rho}i}$ that define $\mathcal{P}{Y|X}(\mathbf{y};{\hat{\mathbf{y}}_i,\hat{\nu}i\ldots\hat{\rho}i})$, optimizing these through likelihood maximization. The distribution $\mathcal{P}{Y|X}$ is chosen to allow uncertainty estimation via a closed-form solution, dependent on the network's parameters.

As Figure \ref{fig:arch} shows, we employ a Bayesian neural network, denoted as $\Omega(\cdot;\theta)$, to predict the per-pixel mean $(\hat{y}_{i})$ alongside the uncertainty parameters scale $(\hat{\alpha}_{i})$ and shape $(\hat{\beta}_i)$ that characterize the per-pixel uncertainty. The scale parameter $(\hat{\alpha}_{i})$ indicates the expected deviation of each prediction, while the shape parameter $(\hat{\beta}_i)$ adapts the distribution's tails to the presence of outliers or other irregularities, thereby capturing the heteroscedastic nature of the data. As they are all trainable parameters($i.e.$ $\{\hat{\mathbf{y}}_{i},\hat{\alpha}_{i},\hat{\beta}_{i}\}:=\Omega(\mathbf{x}_{i};\theta),$), we can describe the optimization problem as the following equation:
\begin{equation}
\begin{aligned}
\theta^*:&={\mathrm{argmax}} \prod_{i=1}^{N}\mathcal{P}_{Y|X}(\mathbf{y}_{i};\{\hat{\mathbf{y}}_{i},\hat{\alpha}_{i},\hat{\beta}_{i}\})
\\&={\mathrm{argmax}}\prod_{i=1}^N\frac{\hat{\beta}_i}{2\hat{\alpha}_i\Gamma(\frac{1}{\hat{\beta}_i})}e^{-(|\hat{\mathbf{y}}_i-\mathbf{y}_i|/\hat{\alpha}_i)^{\hat{\beta}_i}}
\\&={\operatorname*{argmin}}\sum_{i=1}^{N}\left(\frac{|\hat{\mathbf{y}}_{i}-\mathbf{y}_{i}|}{\hat{\alpha}_{i}}\right)^{\hat{\beta}_{i}}-\log\frac{\hat{\beta}_{i}}{\hat{\alpha}_{i}}+\log\Gamma(\frac{1}{\hat{\beta}_{i}})
\end{aligned}
\end{equation}

After training the Bayesian model, we can set the the predicted variance itself as uncertainty in the prediction, $i.e.$, the uncertainty mask $M_{Bayes}$ that we need to use for our subsequent processes:
\begin{equation}
M_{Bayes}=\frac{\hat{\alpha}_i^2\Gamma(\frac3{\hat{\beta}_i})}{\Gamma(\frac1{\hat{\beta}_i})}
\end{equation}

\textbf{Refinement of Uncertainty Masks Using Nonlinear Transformation.} Upon obtaining the initial Bayesian uncertainty mask $M_{Bayes}$ , our method applies a nonlinear transformation to refine this mask, creating a more discriminating variable, denoted $B$. This refined mask $B$ is critical for modulating the noise profile in the subsequent diffusion process, allowing for a tailored noise addition that is more attuned to the nuances of image content and uncertainty. This transformation hinges on a sigmoidal function, traditionally used in neural networks to introduce nonlinearity. The sigmoid function is defined as $\sigma(x)=\frac1{1+e^{-x}}$, mapping real-valued inputs into a (0, 1) range, providing a smooth transition between two states. In the context of our model, the sigmoid function is employed to differentiate between regions that require noise amplification versus those that need noise suppression within the super-resolution process. For each pixel indexed by $i$, the adjustment factor $A_{i}$ is computed using the sigmoid function applied to the mask $M_{Bayes}$ subtracted by a threshold $\alpha$ , and scaled to ensure a steep transition around this threshold. Mathematically, the adjustment factor is given by: $A_i=\sigma((M_{Bayes,i}-\alpha)\cdot k)$, where $k$  is a scaling constant set to provide a sensitive response to the deviations from $\alpha$, which is typically chosen as 10 to amplify the effect.

The amplification and reduction factors for each pixel, are then adjusted according to the calculated adjustment factor based on predefined base values amplification base $\delta_{1}$ and reduction base $\delta_{2}$, These factors ensure that areas with uncertainty levels above $\alpha$ undergo noise amplification, while regions below this threshold experience noise reduction. The final transformation applied to $M_{Bayes}$ to produce the refined mask $B$ is as follows:
\begin{equation}
B=\begin{cases}N_{i}\cdot[(\delta_{1}+(A_{i}-0.5)\cdot\gamma)] & \text{if }M_{Bayes,i}>\alpha\\ N_{i}\cdot[(\delta_{2}-(0.5-A_{i})\cdot\gamma)] & \text{if }M_{Bayes,i}\leq\alpha\end{cases}
\end{equation}
Here, \( N_i \) is the Gaussian noise at the i-th pixel, \( A_i \) is the sigmoid-transformed adjustment factor for the i-th pixel, \( \delta \) is the base value for noise modulation, and \( \gamma \) is the amplification or reduction intensity which scales the contribution of the adjustment factor. This formulation allows for a unified base value \( \delta \) to be modulated by an intensity factor \( \gamma \), enhancing the model's flexibility to calibrate noise levels effectively across the image.
    
\textbf{Integration of Uncertainty in the Diffusion Process.}
In our framework, we use a Bayesian refinement mask $B$ to modulate the heteroscedastic generalized Gaussian noise used in the original SRDiff by applying $B$ to ${\boldsymbol{\epsilon}}_t$. Then the sampling of $x_t$ becomes:
\begin{equation}
\label{equ:xt}
\boldsymbol{x}_t=\sqrt{1-\beta_t}\boldsymbol{x}_{t-1}+\sqrt{\beta_t}(\boldsymbol{\epsilon}_t \odot {B})
\end{equation}
Let $\alpha_t=1-\beta_t$ and iteratively apply Equation \ref{equ:xt}, it comes:

\begin{equation}
x_{t}(x_{0},\epsilon)=\sqrt{\bar{\alpha}_{t}}x_{0}+\sqrt{1-\bar{\alpha}_{t}}(\epsilon \odot B),\epsilon\sim\mathcal{N}(\mathbf{0},\mathbf{I}).
\end{equation}

where $\bar{\alpha}_t=\prod_{i=1}^t\alpha_i$, Our novel diffusion equation modulates noise variance at each pixel according to the corresponding uncertainty:
\begin{equation}
q(x_t|x_0,B)=\mathcal{N}\left(x_t;\sqrt{\bar{\alpha}_t}x_0,((1-\bar{\alpha}_t)B^2I)\right)
\end{equation}

To achieve the SR image from restoration of an LR image, learning the reverse of the forward diffusion process is essential, characterized by the posterior distribution $p(\bm{x}_{t-1}|\bm{x}_t,B)$. However, the intractability arises due to the known marginal distributions $p(\bm{x}_{t-1})$ and $p(\bm{x}_{t})$.This challenge is addressed by incorporating $\bm{x}_0$ into the condition.
Employing Bayes' theorem, the posterior distribution $p(\bm{x}_{t1}|\bm{x}_t,\bm{x}_0,\text{B})$ can be formulated as:
\begin{small}
\begin{equation}
    \begin{aligned}
        & \tilde{\mu}(x_t,t,B):=\frac1{\sqrt{\alpha_t}}\left(x_t-\frac{\beta_t}{\sqrt{1-\bar{\alpha}_t}}\epsilon_\theta(x_t,t,B)\right), \\
        & \tilde{\beta}_t=\frac{1-\bar{\alpha}_{t-1}}{1-\bar{\alpha}_t}\beta_t, \\
        & p(\bm{x}_{t-1}|\bm{x}_t,\bm{x}_0,B) =\mathcal{N}(\bm{x}_{t-1};\tilde{\mu}_t(\bm{x}_t,\bm{x}_0,B), \tilde{\beta}_t B^2\mathbf{I}), \\
    \end{aligned}
    \label{eq:posterior}
\end{equation}
\end{small}
where $\epsilon_\theta$ is a noise predictor.

\textbf{Training.} In the Training Phase, we train BUFF using LR-HR image pairs from the dataset over $T$ total diffusion steps. The conditional noise predictor $\epsilon_{\theta}$ begins with random initialization, while the RRDB-based LR encoder $D$ is pre-trained using an L1 loss function alongside our Bayesian model $\Omega$. For each training iteration, we select a mini-batch of LR-HR pairs, compute the residual image $x_r$, and process LR images through the Bayesian model to produce a modification mask $B$. This mask is used to modulate Gaussian noise and, after merging with LR, is encoded and input into $\epsilon_{\theta}$ with $t$ and $x_t$. We modulate sampled Gaussian noise for each $t$ within ${1,\cdots,T}$ and iteratively optimize the noise predictor through gradient steps, streamlining the process while ensuring the training's effectiveness and efficiency.

\begin{table*}[htbp]
    \centering
    \small

    \resizebox{\textwidth}{!}{
    \begin{tabular}{l|cc|cc|cc|cc|cc|cc}
        \toprule
        & \multicolumn{2}{c|}{\textbf{Set14}} & \multicolumn{2}{c|}{\textbf{Urban100}} & \multicolumn{2}{c|}{\textbf{BSD100}} & \multicolumn{2}{c|}{\textbf{Manga109}} & \multicolumn{2}{c|}{\textbf{General100}} & \multicolumn{2}{c}{\textbf{DIV2K}}                                                                                                                                                                                             \\
        \cmidrule(){2-13}
        & PSNR           & SSIM                             & PSNR           & SSIM                          & PSNR           & SSIM                        & PSNR           & SSIM                         & PSNR           & SSIM                      & PSNR           & SSIM                   \\
        \multirow{-3}*{\textbf{Method}}                                                   & ($\uparrow$)   & ($\uparrow$)             & ($\uparrow$)                           & ($\uparrow$)                                                 & ($\uparrow$) & ($\uparrow$)  & ($\uparrow$) & ($\uparrow$)  & ($\uparrow$) & ($\uparrow$)  & ($\uparrow$) & ($\uparrow$)  \\
        \midrule
        SRCNN                                                  & 24.45          & 0.6432                    & 21.95          & 0.6457                                & 24.69        & 0.6365          & 20.72        & 0.7008         & 22.19        & 0.6432                 & 24.70        & 0.6929         \\
        EDSR\                                                      & 26.72          & 0.7428                    & 23.49          & 0.7233                               & 25.78        & 0.6808         & 29.42        & 0.8798        & 27.25        & 0.7886                & 29.29        & 0.8027       \\
        RCAN                                                       & 26.81          & 0.7440                   & 23.56          & 0.7241                    & 25.69        & 0.6797        & 29.09        & 0.8746        & 27.18        & 0.7861               & 29.32        & 0.8033       \\
        ESRGAN                                                              & 25.27          & 0.6801                    & 22.99          & 0.6940              & 24.65        & 0.6374         & 28.60        & 0.8553         & 26.03        & 0.7449               & 27.18        & 0.7709         \\
        Real-ESRGAN                                                   & 25.18          & 0.7098                     & 22.12          & 0.6869                                     & 25.11        & 0.6712         & 26.73        & 0.8639               & 25.64        & 0.7607               & 27.56        & 0.7893       \\
        BSRGAN~                                                  & 25.05          & 0.6746                          & 22.37          & 0.6628                   & 24.95        & 0.6365               & 26.09        & 0.8272               & 25.23        & 0.7309                & 27.32        & 0.7577              \\
        BebyGAN                                                         & 25.73          & 0.6994                  & 23.36          & 0.7113                               & 24.75        & 0.6527        & 29.35        & 0.8775        & 26.19        & 0.7549                 & 28.62        & 0.7904       \\
        SwinIR                                                       & 26.77          & 0.7269                    & 25.06          & 0.7488                               & 26.11        & 0.6913         & 28.94        & 0.8687        & 27.83        & 0.8015                & 28.19        & 0.7727        \\
        HAT                                                       & 27.09          & 0.7482                    & 25.38          & 0.7642                               & 26.41        & 0.6998         & 29.33        & 0.8839        & 30.18        & 0.8297                & 29.21        & 0.8066        \\
        IDM                                                 & 26.53          & 0.7255                  & 24.87          & 0.7479                              & 26.04        & 0.6938         & 29.11        & 0.8697        & 27.79       & 0.8005                & 28.46        & 0.7898        \\
        ACDMSR                                                       & 26.81          & 0.7398                   & 25.15          & 0.7589                               & 25.98        & 0.6875         & 29.21        & 0.8842       & 30.24        & 0.8337                & 28.87        & 0.7956        \\
        SAM-DiffSR                                                & 27.01          & 0.7456                   & 25.46          & 0.7621                             & 26.39        & 0.7003         & 29.36        & 0.8899        & 30.06        & 0.8353                 & 28.84        & 0.8009        \\
        ResDiff                                                   & 26.73          & 0.7457                    & 25.21          & 0.7629                              & 26.32        & 0.6951         & 29.23        & 0.8739       & 30.11        & 0.8241                 & 28.77        & 0.8023       \\
        
        \midrule
        SRDiff(Baseline)                                                              & 26.69          & 0.7287                   & 25.13          & 0.7582             & 25.79        & 0.6813         & 28.82        & 0.8725       & 29.79        & 0.8241                & 28.76        & 0.7912        \\
        
        \rowcolor[gray]{0.85}
        \cellcolor{white}BUFF (Ours)                                                & 27.11          & 0.7487                   & 25.49          & 0.7634                               & 26.40        & 0.7011         & 29.38        & 0.8861        & 30.21        & 0.8349                & 29.35        & 0.8078      \\
        
        \rowcolor[gray]{0.85}
        \cellcolor{white}\enspace\enspace\enspace\enspace\enspace\enspace\enspace$\Delta$         & \textbf{+0.42}                                  & \textbf{+0.0200}                                                      & \textbf{+0.36}        & \textbf{+0.0052}         & \textbf{+0.61}        & \textbf{+0.0198}        & \textbf{+0.56}        & \textbf{+0.0136}                 & \textbf{+0.42}        & \textbf{+0.0108}         & \textbf{+0.59}        & \textbf{+0.0166}        \\
        \bottomrule
    \end{tabular}
    }
    \caption{Results on test sets of several public benchmarks and the validation set of DIV2K. The first 12 rows report the results achieved by MSE-based, GAN-based, Flow-based and diffusion-based approaches. $\Delta$ represents performance improvements over the diffusion-based baseline SRDiff. ($\uparrow$) and ($\downarrow$) indicate that a larger or smaller corresponding score is better, respectively.}
    \label{tab:main}
\end{table*}

\textbf{Inference.} A $T$-step BUFF inference begins by taking a low-resolution (LR) image $x_L$ as input. Initially, we draw a latent variable $x_T$ from a standard Gaussian distribution and upscale $x_L$ using a bicubic kernel. Simultaneously, we employ a Bayesian model to generate a corresponding mask $B$, mirroring the training process. Moreover, the LR image $x_L$ and mask $B$ are encoded into $x_e$ by the LR encoder—a step executed just once before iteration commencement—to expedite the inference process. Iterations initiate from $t = T$, with each iteration yielding a residual image characterized by progressively diminishing noise levels as $t$ decreases. For iterations where $t > 1$, Gaussian noise $z$ is sampled and modulated, and $x_{t-1}$ is computed employing the noise predictor $\epsilon_\theta$ with inputs $x_t$, $x_e$, and $t$. Upon reaching $t = 1$, we set $z = 0$, and $x_0$ emerges as the final residual prediction. The super-resolution (SR) image is reconstructed by adding the residual image $x_0$ to the upscaled LR image, denoted as up($x_L$).

\section{Experiment}
\subsection{Experimental Setup}

\begin{figure*}[htbp]
    \centering
    \includegraphics[width=\linewidth]{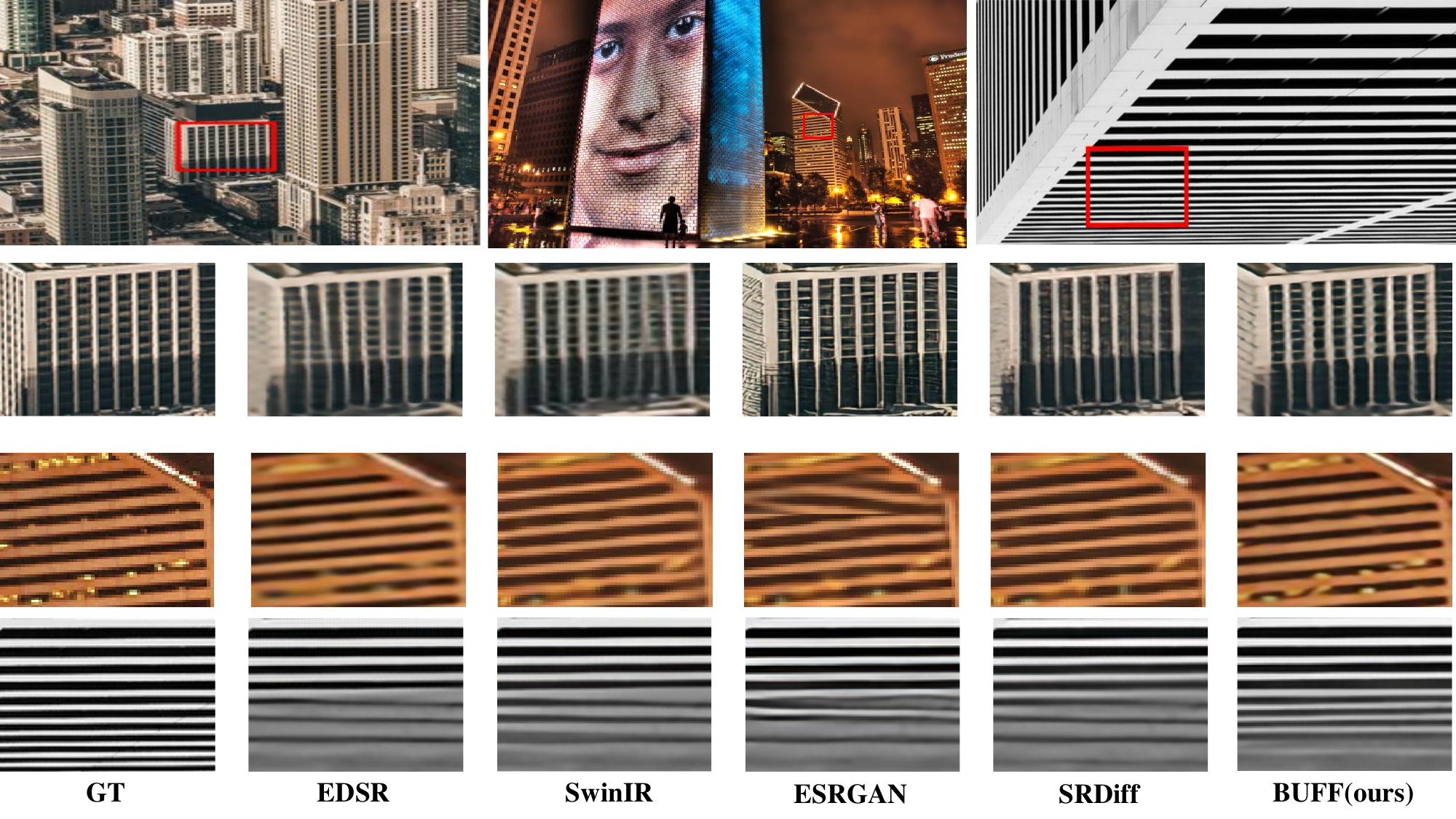}
    \caption{Visualization of restored images generated by different methods. Our BUFF surpasses other approaches in terms of both higher reconstruction quality and fewer artifacts. Additional visualization results can be found in our supplementary material.}
    \label{fig:visvis}
    \vspace{-2mm}
\end{figure*}

\textbf{Datasets and benchmarks.} We train both the Bayesian model and BUFF on the DF2K dataset, a combination of DIV2K \cite{DIV2K} and Flickr2K, comprising 3450 (800 + 2650) high-quality images. After Bayesian model's training, for all images in the training set, we adopt the Bayesian model to obtain their corresponding  masks. We then adopt a patch size settings of $160\times160$ to crop each image and its corresponding mask. During testing, we evaluate our models using PSNR and SSIM on six publicly available benchmark datasets: Set5 \cite{Set5}, Set14 \cite{Set14}, B100 \cite{B100}, Urban100 \cite{Urban100}, Manga109 \cite{Manga109} and DIV2K \cite{DIV2K}. For face SR, we train the models at $16\times16 \to 128 \times 128$ on Flickr-Faces-HQ (FFHQ) dataset, which includes 70k images in total, and we sample 400 images from CelebA-HQ dataset for evaluation. Both objective and subjective metrics are used in our experiment. To evaluate the perceptual quality, we also adopt Frechet inception distance (FID) \cite{fid} as the subjective metric, which measures the fidelity and diversity of generated images. PSNR and SSIM results are calculated on the \textbf{Y} channel in the YCbCr color space.

\textbf{Training implementation details.} We trained the Bayesian model for 50k rounds using a batchsize of 16, and the training strategy uses the NLL loss mentioned in Section.\ref{sec:method}, and the initial learning rate of \( \eta = 1 \times 10^{-4} \), which was decayed by a factor of 0.5 every \( 2 \times 10^5 \) iterations. We employed the Adam optimizer with \( \beta_1 = 0.9 \) and \( \beta_2 = 0.999 \), without weight decay. We train the diffusion model for 400K iterations with a batch size of 16, and adopt Adam as the optimizer. The initial learning rate is \( \eta = 2 \times 10^{-4} \) and the cosine learning rate decay is adopted. All experiments were conducted on a system equipped with an NVIDIA RTX 3090 GPU, and the models were implemented using the PyTorch framework.

\subsection{Performance of Image SR}
Our proposed BUFF method showcases a remarkable performance. Table \ref{tab:main}, which compiles our extensive benchmarking results, indicates that BUFF surpasses the diffusion-based baseline SRDiff across nearly all metrics. Notably, BUFF achieves impressive enhancements in PSNR and SSIM across multiple datasets. While there's a marginal decrease in the SSIM on Set14 and BSD100, this is offset by significant gains in image fidelity elsewhere. Visual evidence of BUFF's superior reconstruction capabilities is provided in Figure \ref{fig:visvis}. Here, BUFF-generated images exhibit notably sharper details and more faithful textures when compared to baseline methods. The comparison highlights BUFF's ability to reduce artifacts, those unwanted distortions that often accompany SR processes, thereby ensuring a cleaner and more accurate rendition of the original scene. 

\begin{figure}[htbp]
    \centering
    \includegraphics[width=0.9\linewidth]{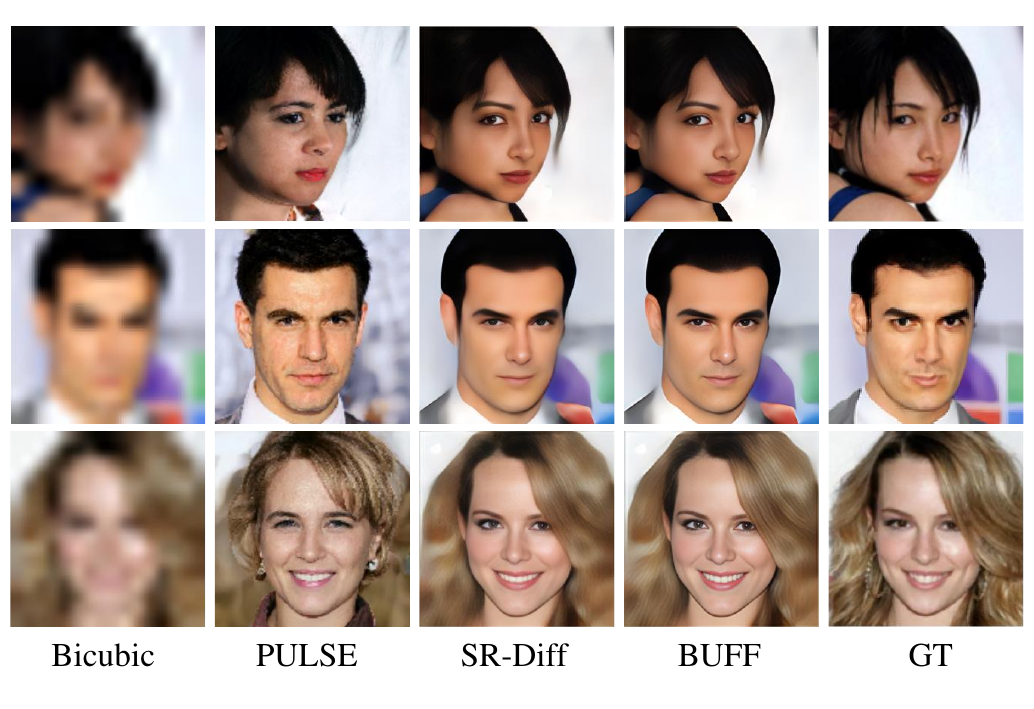}
    \caption{Visual comparisons on CelebA-HQ dataset for 8×
face SR. Zoom in for a better view.}
    \label{fig:face}
    \vspace{-4mm}
\end{figure}  

\begin{table}[htbp]
\centering
\small
\begin{tabularx}{\linewidth}{l
  *{6}{>{\centering\arraybackslash}X}
}
\toprule
\textbf{Method} & \multicolumn{3}{c}{\textbf{DIV2K100}} & \multicolumn{3}{c}{\textbf{ImageNet-1K}} \\
\cmidrule(lr){2-4} \cmidrule(lr){5-7}
 & {\textbf{LPIPS} } & {\textbf{FID} } & {\textbf{PSNR} } & {\textbf{LPIPS} } & {\textbf{FID} } & {\textbf{PSNR}} \\
\midrule
DASR {}      & 0.4476 & 149.11 & 25.46 & 0.4116 & 100.66 & 26.22 \\
DAN {}       & 0.3597 & 96.63  & 26.74 & 0.3272 & 68.52  & 27.33 \\
DCLS {}      & 0.3085 & 69.98  & 28.31 & 0.2791 & 54.59  & 29.02 \\
BSRGAN {}    & 0.3526 & 98.39  & 24.90 & 0.3546 & 80.95  & 25.60 \\
AdaTarget {} & 0.2923 & 77.04  & 28.25 & 0.3249 & 56.81  & 27.58 \\
DARSR {}     & 0.4956 & 148.34 & 24.05 & 0.4618 & 107.79 & 24.22 \\
KDSR {}      & 0.4328 & 144.25 & 25.82 & 0.4035 & 101.22 & 26.48 \\
\rowcolor{gray!25}
SRDiff    & 0.3041 & 56.12  & 26.63 & 0.2772 & 64.12  & 27.41 \\
\rowcolor{gray!25}
BUFF (ours)   & 0.2946 & 51.62  & 26.75 & 0.2741 & 52.42  & 28.29 \\
\bottomrule
\end{tabularx}
\caption{Quantitative results on DIV2K100 and ImageNet-1K datasets.}
\label{tab:my_label}
\vspace{-10mm}
\end{table}

% There's a well-known perception-distortion trade-off, suggesting that enhancing a model's generative power - whether by increasing diffusion steps or boosting adversarial loss in GANs—tends to reduce fidelity even as it improves image authenticity. This is because more capable models often add high-frequency details that stray from the truth. To compare our BUFF with SRDiff, we illustrated their perception-distortion tradeoffs in Figure.\ref{trade}, using LPIPS and mean square-error (MSE) for perception and distortion measures. The plot shows BUFF and SRDiff's performance over various diffusion steps (10, 15, 20, 30, and 40), highlighting BUFF's better balance between perception quality and reconstruction fidelity, as evidenced by its consistently lower curve compared to SRDiff's.

\subsection{Performance of Deblurring}
In an effort to benchmark the deblurring efficacy of our BUFF, we subjected it to a series of evaluations against several state-of-the-art methodologies on the DIV2K100 and ImageNet-1K datasets. The high-resolution (HR) images were synthetically degraded using random anisotropic Gaussian kernels to simulate a range of blur conditions. BUFF's performance was measured against seven contemporary deblurring methods—ranging from DASR\cite{dasr}, DAN\cite{dan}, and DCLS\cite{dcls}, to BSRGAN\cite{bsrgan}, AdaTarget\cite{AdaTarget}, DARSR\cite{DARSR}, and KDSR\cite{KDSR}. Quantitative results presented in Table \ref{tab:my_label} reveal BUFF's superior performance, even amidst the generalized and demanding conditions of the tests. These improvements prove that BUFF also has a very great advantage and potential handling complex deblurring tasks compared to SRDiff and general deblurring models.

\subsection{Performance of Face SR}
Figure \ref{fig:face} showcases the performance of our model on the CelebA-HQ dataset for $8\times$ face super-resolution (SR). The comparison spans several methods: bicubic interpolation, PULSE\cite{pulse}, SRDiff, and our BUFF model, against the high-resolution (HR) benchmarks. Our BUFF model exhibits a remarkable improvement in recovering fine details and producing lifelike textures, as seen in the clarity of facial features. This visual assessment underscores BUFF's advanced capabilities in enhancing image quality, affirming its effectiveness in generating high-fidelity face SR images. 

\subsection{Ablation study}
\label{sec:ablation}

\begin{figure}[htbp]
  
  \centering
  \includegraphics[width=\linewidth,height=0.55\linewidth]{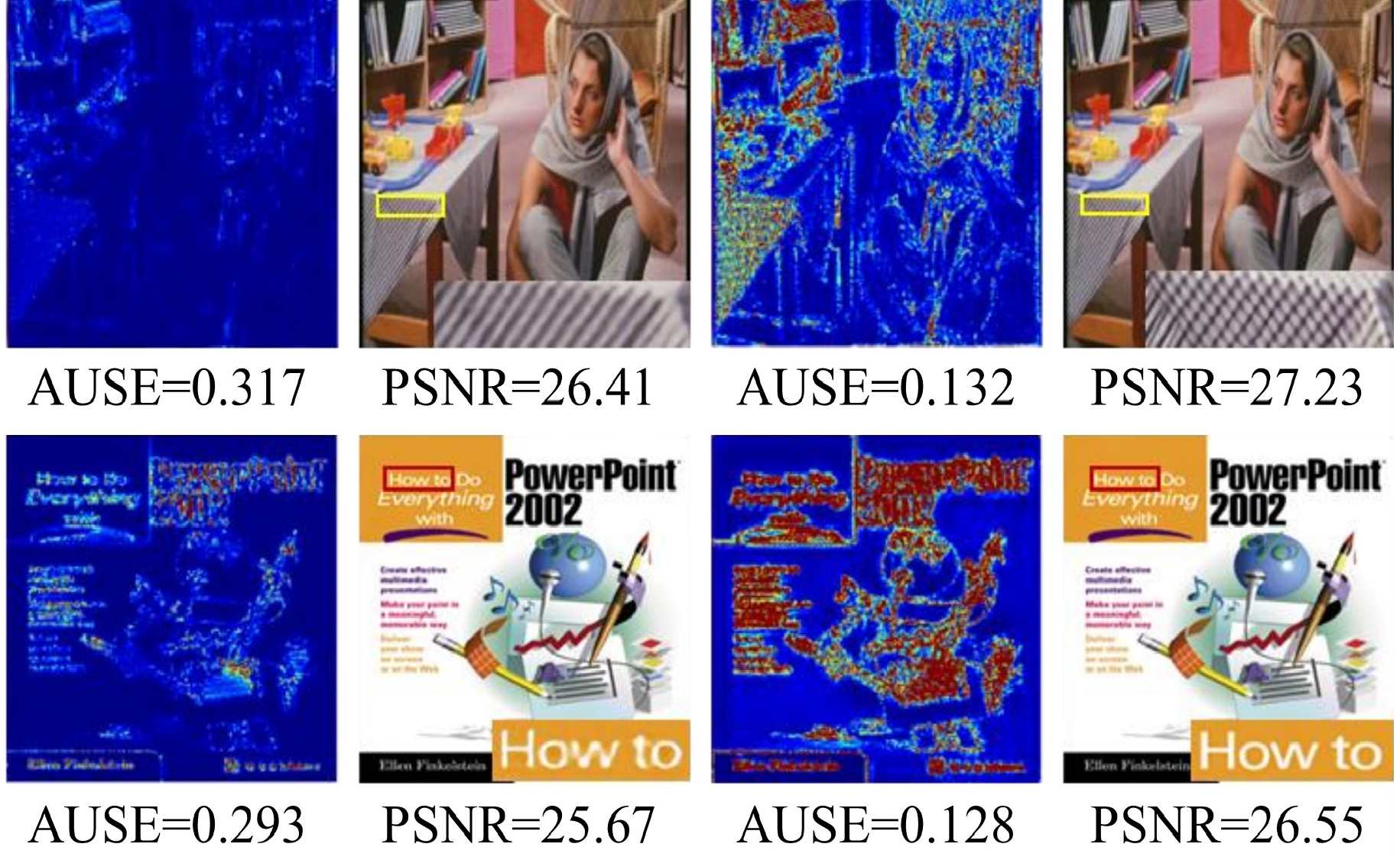}
  \vspace{-5mm}
  \caption{Visualization of the effect of uncertainty Mask Quality.}
  \label{fig:ablation2}
  \vspace{-4mm}
\end{figure}

{\bf Quality of Uncertainty mask.} In our BUFF study, we examined how uncertainty masks of different qualities, produced by Bayesian models at varying training levels, affect super-resolution performance. From Table.\ref{tab:ablation_mask}, these qualities, reflected by Area Under the Sparsification Error (AUSE) scores (0.10, 0.20, 0.30), relate to the precision of uncertainty estimation—the lower the AUSE, the better the estimation. We categorized the masks into "Low," "Medium," and "High" qualities based on their AUSE scores and integrated each into the diffusion model. Evaluating the model on Urban100 and DIV2K datasets revealed a clear pattern: better mask quality leads to improved model performance, evidenced by metrics like PSNR, SSIM, and FID. This highlights the crucial role of accurate uncertainty estimation in super-resolution, with high-quality masks enhancing image fidelity and detail by aligning noise modulation with actual errors more effectively. We also present the visual results of the ablation experiment in Fig. \ref{fig:ablation2}, which verifies the effect of uncertainty masks on super-resolution output.

\begin{table}
    \vspace{2.3mm}
    \renewcommand\tabcolsep{4pt}
    \centering
    \small
    
    \vspace{-2mm}
    \scalebox{1}{
        \begin{tabular}{c|ccc|ccc}
            \toprule
            & \multicolumn{3}{c|}{\textbf{Urban100}} & \multicolumn{3}{c}{\textbf{DIV2K}} \\
            \cmidrule(){2-7}
            & PSNR           & SSIM            & FID             & PSNR           & SSIM            & FID             \\
            \multirow{-3}*{\textbf{\makecell{Mask \\quality\\(AUSE)}}}    & ($\uparrow$)          & ($\uparrow$)          & ($\downarrow$)          & ($\uparrow$)          & ($\uparrow$)          & ($\downarrow$)          \\
            \midrule
            0.308& 25.01          & 0.7582          & 4.93          & 29.01          & 0.7991          & 0.45          \\
            0.217   & 25.23 & 0.7605 & 4.87 & 29.12 & 0.8013 & 0.44 \\
            0.121   & \textbf{25.49} & \textbf{0.7634} & \textbf{4.61} & \textbf{29.35} & \textbf{0.8078} & \textbf{0.41} \\
            \bottomrule
        \end{tabular}
    }
    \caption{Comparison of uncertainty masks with different qualities.}
    \label{tab:ablation_mask}
\end{table}

\begin{table}
    \renewcommand\tabcolsep{3.2pt}
    \centering
    \small
    
    \vspace{-2mm}
    \scalebox{1}{
        \begin{tabular}{c|ccc|ccc}
            \toprule
            & \multicolumn{3}{c|}{\textbf{Urban100}} & \multicolumn{3}{c}{\textbf{DIV2K}} \\
            \cmidrule(){2-7}
            & PSNR           & SSIM            & FID             & PSNR           & SSIM            & FID             \\
            \multirow{-3}*{\textbf{\makecell{Amplification\\Intensity}}} & ($\uparrow$)          & ($\uparrow$)          & ($\downarrow$)          & ($\uparrow$)          & ($\uparrow$)          & ($\downarrow$)          \\
            \midrule
            1.30 & 25.41          & 0.7631          & 4.60          & 29.28          & 0.8070          & 0.45         \\
            1.10   & 25.32 & 0.7631 & 4.64 & 29.17 & 0.8045 & 0.47 \\
            1.20(ours)   & \textbf{25.49} & \textbf{0.7634} & \textbf{4.61} & \textbf{29.35} & \textbf{0.8078} & \textbf{0.41} \\
            \bottomrule
        \end{tabular}
    }
    \caption{Comparison of different intensity for noise amplification.}
    \label{tab:ablation_spe}
    \vspace{-4mm}
\end{table}

{\bf Comparison of Amplification Intensity.}
Our investigation into noise amplification within super-resolution tasks in Table.\ref{tab:ablation_spe} shows that overly high amplification factors ($e.g.$, 1.30) adversely affect both training efficiency and image quality, as demonstrated by performance drops in the Urban100 and DIV2K datasets. This indicates that excessive noise can destabilize the super-resolution model, complicating training and diminishing output quality. On the other hand, lower amplification levels, like 1.10, fail to significantly improve performance and introduce unnecessary complexity compared to the baseline SRDiff model. An amplification setting of 1.20, however, finds a sweet spot by improving key metrics such as PSNR and SSIM, and by reducing the FID.

\begin{table}[htbp]

\centering
\begin{adjustbox}{width=\linewidth}
\begin{tabular}{c|ccccccc}
\hline % Adjust the shade of gray as needed
Method & BG & BE & PSNR$\uparrow$ & SSIM$\uparrow$ & DI$\uparrow$ & LPIPS$\downarrow$ \\
\hline
a& \checkmark &  & 26.21 & 0.7004 & 15.31 & 0.2834 \\
b& & \checkmark  & 26.01 & 0.6992 & 15.15 & 0.2958 \\
\rowcolor{gray!30}
c& \checkmark &\checkmark & \textbf{26.40} & \textbf{0.7011} & \textbf{15.36} & \textbf{0.2741} \\
\hline
\end{tabular}
\end{adjustbox}
\caption{Comparison of model configurations and their performance metrics for $\times4$ SR task on BSD100 dataset.}
\label{tab:model_configurations}
\vspace{-2mm}
\end{table}

{\bf Directly integrating Mask into diffusion model.}
In our study, we performed an ablation test on the BSD100 dataset for a $4\times$ super-resolution (SR) task to assess the impact of Bayesian Guided (BG) and Bayesian Embedding (BE) components in our SR model depicted in Table.\ref{tab:model_configurations}. The baseline 'a' applies the BG method, using Bayesian-generated masks for noise modulation, while 'b' employs BE, adding refined masks to the noise predictor with encoded low-resolution inputs. Model 'c' combines BG and BE, enhancing both noise modulation and prediction accuracy. Results show BG and BE independently improve PSNR and SSIM metrics, with BUFF achieving the best scores across all metrics, including the lowest LPIPS, indicating superior image quality. 

\vspace{-2mm}
\section{Conclusion}
In this work, we have presented BUFF, a novel framework that augments diffusion-based image super-resolution models by integrating Bayesian-derived uncertainty masks to refine structure-level detail enhancement. Our method effectively injects structural information into the diffusion process, tuning the noise profile at a pixel-level based on localized uncertainty. This targeted modulation of noise leads to a marked improvement in the delineation of structural details and a concurrent reduction in image artifacts. The performance of our approach has been rigorously validated through comprehensive testing on standard image super-resolution benchmarks, confirming its efficacy and potential applicability in advanced imaging tasks.

\section{Acknowledgments}
This work was supported by National Science and Technology Major Project (No. 2022ZD0118201), the National Science Fund for Distinguished Young Scholars (No.62025603), the National Natural Science Foundation of China (No. U21B2037, No. U22B2051, No. U23A20383, No. U21A20472, No. 62176222, No. 62176223, No. 62176226, No. 62072386, No. 62072387, No. 62072389, No. 62002305 and No. 62272401), and the Natural Science Foundation of Fujian Province of China (No. 2021J06003, No.2022J06001).

\bigskip

\bibliography{aaai25}

\end{document}